\documentclass[orivec,runningheads]{llncs}
\usepackage[T1]{fontenc}
\usepackage{graphicx}
\usepackage{hyperref}
\hypersetup{
    colorlinks=true,
    linkcolor=blue, 
    filecolor=blue,
    urlcolor=blue,
    citecolor=blue
}

\usepackage{amsmath}
\usepackage{algorithm}
\usepackage{algpseudocode}
\usepackage{bbm}
\usepackage{booktabs}
\usepackage{MnSymbol}
\usepackage{wrapfig}
\usepackage{mathbbol}
\usepackage{marvosym}
\begin{document}
\title{SAU: A Dual-Branch Network to Enhance Long-Tailed Recognition via Generative Models}
\titlerunning{Enhance Long-Tailed Recognition via Generative Models}
%
\author{Guangxi Li \and
Yinsheng Song \and
Mingkai Zheng\textsuperscript{\Letter}}
\authorrunning{G. Li et al.}
%
\institute{School of Computer Science, Faculty of Engineering, The University of Sydney
\email{\{guli5858, yson6207\}@uni.sydney.edu.au} \\ \email{mingkaizheng@outlook.com}}
\maketitle              
\begin{abstract}
Long-tailed distributions in image recognition pose a considerable challenge due to the severe imbalance between a few dominant classes with numerous examples and many minority classes with few samples. Recently, the use of large generative models to create synthetic data for image classification has been realized, but utilizing synthetic data to address the challenge of long-tailed recognition remains relatively unexplored. In this work, we proposed the use of synthetic data as a complement to long-tailed datasets to eliminate the impact of data imbalance. To tackle this real-synthetic mixed dataset, we designed a two-branch model that contains Synthetic-Aware and Unaware branches (SAU). The core ideas are (1) a synthetic-unaware branch for classification that mixes real and synthetic data and treats all data equally without distinguishing between them. (2) A synthetic-aware branch for improving the robustness of the feature extractor by distinguishing between real and synthetic data and learning their discrepancies. Extensive experimental results demonstrate that our method can improve the accuracy of long-tailed image recognition. Notably, our approach achieves state-of-the-art Top-1 accuracy and significantly surpasses other methods on CIFAR-10-LT and CIFAR-100-LT datasets across various imbalance factors. Our code is available at \href{https://github.com/lgX1123/gm4lt}{{https://github.com/lgX1123/gm4lt}}.

\end{abstract}
\section{Introduction}
In the field of computer vision, mainstream datasets \cite{deng2009imagenet,krizhevsky2009learning} are characteristically balanced. This equilibrium is a crucial factor underpinning the success of deep neural networks in image recognition tasks. Conversely, in real-world scenarios, data frequently exhibits a long-tailed distribution. As a result, models trained on such imbalanced datasets often disproportionately emphasize the head classes and neglect the tail classes, leading to suboptimal performance on a balanced test set. Thus, enhancing model performance in the context of long-tailed distribution data has emerged as a significant challenge.

To mitigate the imbalanced data issue, traditional methods to address data imbalance include class re-balancing strategies such as re-sampling \cite{zhang2021learning} and re-weighting \cite{cui2019class,ren2020balanced}, which aim to correct the imbalance by giving more emphasis to under-represented classes. More recently, diverse approaches have emerged. Contrastive learning \cite{cui2021parametric,wang2021contrastive,zhu2022balanced} enhances the model's feature extraction capabilities and improves the model's ability to differentiate between similar and dissimilar images by comparing. Nevertheless, most of these methods attempt to calibrate the discrimination of the tail classes rather than addressing the root issue of data imbalance.

In recent years, generative models \cite{floridi2020gpt,achiam2023gpt,touvron2023llama,ramesh2021zero,ramesh2022hierarchical,nichol2021glide} have experienced significant advancements, demonstrating immense potential across various applications. Researchers have begun to leverage synthetic data generated by these text-to-image (T2I) models for visual tasks  \cite{he2022synthetic,besnier2020dataset,jahanian2021generative}. Despite their innovative capabilities, synthetic images often suffer from unrealistic hallucinations, which can mislead downstream image processing tasks. Therefore, effectively utilizing the vast amounts of synthetic data generated by T2I models remains a crucial challenge that needs to be addressed.

In this work, inspired by \cite{he2022synthetic,zhao2024ltgc}, we aim to enhance long-tailed image recognition by leveraging powerful large generative models. Specifically, we adopt GLIDE \cite{nichol2021glide} as our text-to-image (T2I) model. To generate diverse and high-quality synthetic images, we utilize GPT-4 \cite{achiam2023gpt} for creating varied image descriptions and employ CLIP \cite{radford2021learning} to filter out low-quality synthetic images. The resulting synthetic data is then used to augment the long-tailed dataset, creating a more balanced dataset for training our models. To effectively handle the real-synthetic mixed dataset, we propose a Synthetic-Aware and Unaware two-branch framework (SAU) to maintain consistency between synthetic-aware and unaware processing. In the synthetic-unaware branch, we apply mix-based augmentations, such as MixUp \cite{zhang2017mixup} and CutMix \cite{yun2019cutmix}, to blend real and synthetic data. This encourages the model to operate without distinguishing between real and synthetic images, making this branch suitable for classification tasks by effectively classifying new images regardless of their origin. In the synthetic-aware branch, we focus on enhancing feature extraction through supervised contrastive learning (SupCon) \cite{khosla2020supervised}. We introduce a K-Nearest Neighbor-based label correction strategy to identify low-quality data within this branch dynamically. Identified low-quality data are treated as noise, and we employ three distinct noise-dropping strategies to design the corresponding objectives.

Our method is mainly evaluated in three widely used public long-tailed image recognition datasets CIFAR-10-LT, CIFAR-100-LT, and ImageNet-LT. Extensive experimental results show our method achieves state-of-the-art Top-1 accuracy and significantly surpasses other methods on CIFAR-10-LT and CIFAR-100-LT datasets over a range of imbalance factors. 

Our main contributions can be summarized as follows:
\begin{itemize}
    \item We propose the utilization of synthetic data generated by Text-to-Image (T2I) models and optionally with LLMs to augment long-tailed datasets, thereby creating a balanced dataset to address data imbalance issues. To effectively manage this dataset, we design a two-branch network architecture comprising Synthetic-Aware and Synthetic-Unaware branches.
    \item We proposed a K-Nearest Neighbor-based label correction procedure to dynamically detect low-quality synthetic data for contrastive learning. We designed three variants of supervised contrastive loss to handle detected low-quality synthetic data. 
    \item The experimental results show that our method achieves state-of-the-art Top-1 accuracy on popular long-tailed benchmarks including CIFAR-10-LT and CIFAR-100-LT.
\end{itemize}

\section{Related Works}
\subsection{Generative Models for Image Recognition}
Recently, generative models have developed rapidly, and they have also obtained great achievements. Specifically, generative models can be divided into two categories, Large Language Models (LLMs) \cite{floridi2020gpt,achiam2023gpt,touvron2023llama} and Text-to-Image (T2I) Models \cite{ramesh2021zero,ramesh2022hierarchical,nichol2021glide}. Additionally, many studies have investigated whether generated contents by these advanced models can help improve the performance of downstream tasks. He et al. \cite{he2022synthetic} show that synthetic data are ready to be applied to image recognition, and synthetic data can significantly enhance the performance of zero-shot and few-shot recognition. Qiu et al. \cite{qiu2021synface} explore the potential of synthetic data for face recognition. However, in the field of long-tailed recognition, there is little research on using generated data to address the problem. An important challenge of long-tailed recognition is the scarcity of tail classes, which can be solved by using generated data.

\subsection{Long-Tailed Recognition}
Existing methods for long-tailed image recognition can be divided into four main parts, including re-sampling, re-weighting, mix-based augmentation, and contrastive learning. Re-sampling \cite{chawla2002smote,zhang2021learning} methods use either over-sampling or under-sampling to create a balanced mini-batch during training. Re-weighting \cite{cui2019class,ren2020balanced} intends to modify the loss function to generate uneven impacts for different classes. Mix-based augmentation \cite{yun2019cutmix,zhang2017mixup} methods enrich tail class representations by generating mixed images with over-sampled tail class images. Contrastive learning \cite{khosla2020supervised,cui2021parametric,zhu2022balanced,du2023global} methods aim to pull the samples of the same class together while pushing samples of different classes apart, which have achieved great success in the field of long-tailed image recognition. However, all of these methods make it difficult to learn the representations of tail classes due to scarcity of data. In this work, by leveraging synthetic data, we aim to use mix-based augmentation methods to reduce the domain gap between real data and synthetic data and use contrastive learning methods to improve the model's understanding and generalization capabilities.

\section{Methods}

\subsection{Overall Framework}
\label{overall_framework}
Our overall framework is shown in Figure \ref{framework}, which can be divided into seven main components:
\begin{figure}[h]
\includegraphics[width=\textwidth]{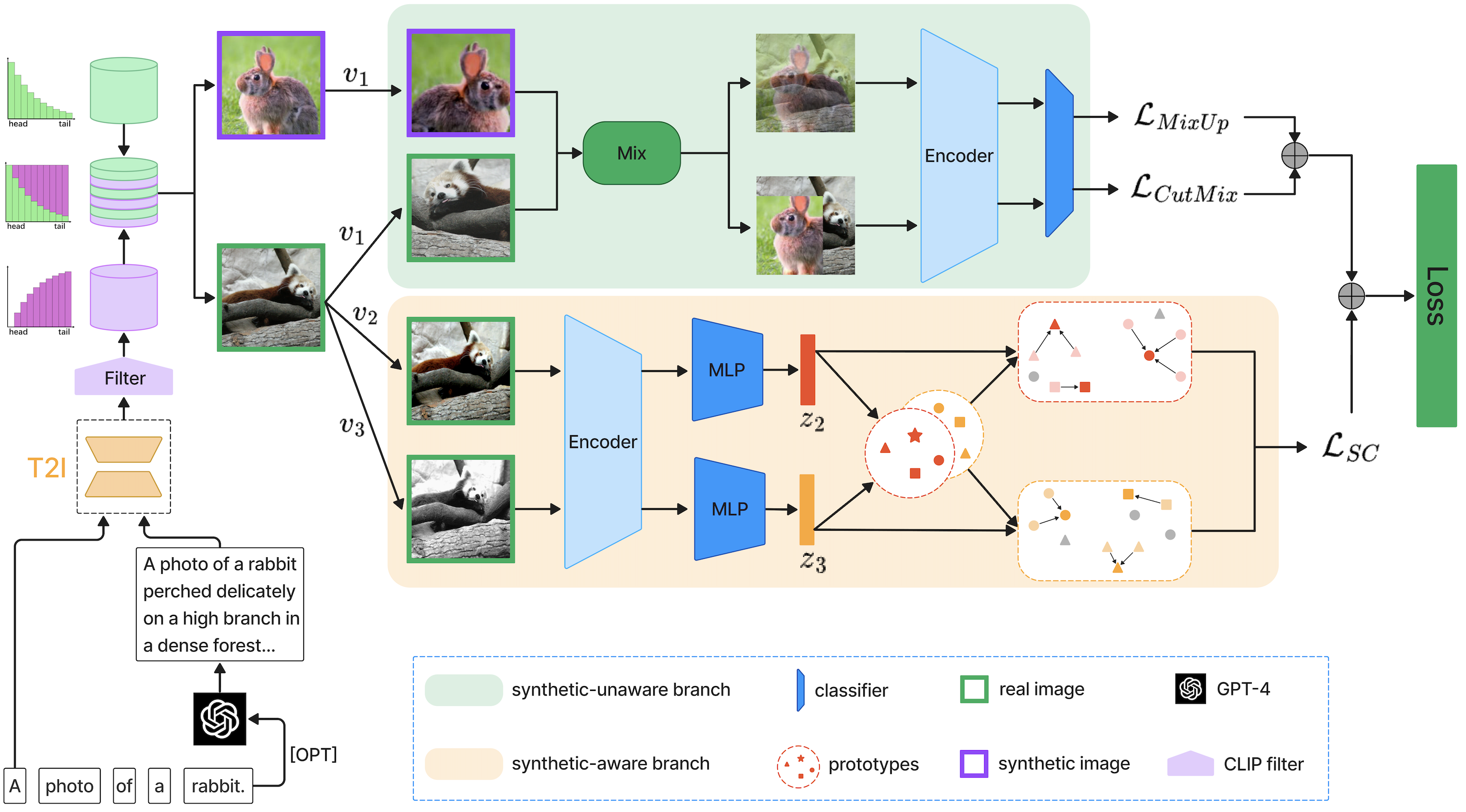}
\caption{The overall framework of our proposed method. By leveraging LLM and T2I models, we generate synthetic data as a complement to a long-tailed distributed dataset to obtain a balanced dataset. Synthetic-unaware branch takes the same view $v_1$ from two samples to calculate the mixing loss. Synthetic-aware branch takes two different views $v_2$ and $v_3$ from one sample to calculate supervised contrastive loss.} \label{framework}
\end{figure}
\begin{itemize}
    \item \textbf{Synthetic data generation module}: For each class that requires synthetic images, we optionally utilize LLMs to enhance the prompts, and then employ T2I models to generate class-related synthetic images.
    \item \textbf{Data augmentation module}: $T(\cdot)$. For each input sample $\mathbf{x}$, we generate three random augmented views, denoted as $\mathbf{v} = T(\mathbf{x})$. Two of these views are utilized in the synthetic-aware branch, while the remaining one is used in the synthetic-unaware branch.
    \item \textbf{Encoder network}: $\mathcal{F}(\cdot)$, which maps input sample $\mathbf{x}$ to feature space, denoted as $\mathbf{h} = \mathcal{F}(\mathbf{x})$. Both the synthetic-unaware branch and synthetic-aware branch share the network.
    \item \textbf{Mix module}: MixUp and CutMix. For each pair of two augmented input samples, it mixes them and their labels into two forms of augmentation pairs. 
    \item \textbf{Classifier head}: A linear classifier head which maps feature vector $\mathbf{h}$ to class space. Note that it calculates the mixed cross-entropy loss during training. 
    \item \textbf{A multi-layer perceptron projection}: $\phi$, which maps feature vector $\mathbf{h}$ to a low-dimensional representations $\mathbf{z}$. 
    \item \textbf{Label correction module}: A powerful module which is used to dynamically detect low-quality synthetic image-label pairs in synthetic-aware branch. 
\end{itemize}

\subsection{Synthetic Data Generation}
\label{3.2}
In our work, we employ LLMs and T2I models to collaboratively generate high-quality and diverse datasets for model training. 

\subsubsection{Prompt with GPT-4.} 
The main purpose of this unit is to create a description of a certain class as a guide prompt for the T2I generation process. Here we have two ways to generate prompts, basic prompts and LLM-enhanced prompts. 

\paragraph{Basic Prompts.}
For the basic prompts, following the simple and effective template introduced by CLIP \cite{radford2021learning}, the fundamental template used for generating descriptions in this context is expressed as:
\begin{equation}
    P_y = \text{"A photo of a [CLASS]."}
\end{equation}
where $P_y$ indicates the prompt for class $y$. However, the main shortcoming of basic prompts is less diverse.

\paragraph{LLM Enhanced Prompts.}
To increase the diversity of the synthetic images, we adopt the powerful tool LLM GPT-4 to generate diverse descriptions for synthetic image generation. Specifically, we assign the GPT-4 model as a role of the system with a system command, such as "\textit{You are a helpful assistant designed to output a prompt (no more than 50 words) describing a given object’s appearance and actions.}". We provide a template of response for a given class $y$ as "\textit{A photo of the class [$y$], {with specific features} and {with a specific background}.}". Last, we also design the query template for a certain class. For example, "\textit{What the [CLASS] looks like?}". In summary, the LLMs enhanced prompts for synthetic image generation can be formulated as: 
\begin{equation}
    {P_y} = \text{GPT-4}_{\text{command}}({\text{Query}_y})
\end{equation}

\subsubsection{T2I with GLIDE.}
\label{t2i}
After obtaining the texts, we employ them to generate synthetic images by leveraging advanced T2I models. In detail, we use GLIDE \cite{nichol2021glide} to generate synthetic images, which can be formulated as:
\begin{equation}
    I^{n}_{y} = \text{GLIDE}(P^{n}_{y})
\end{equation}
where $n \in N_y$ represents the number of synthetic images of a certain class, $I^{n}_{y}$ and $P^{n}_{y}$ indicate the corresponding generated images and prompts for the class $y$, respectively. 

However, some synthetic images might not be high-quality enough to represent the corresponding class. To address this problem, we use CLIP \cite{radford2021learning} to evaluate whether synthetic images are well related to their corresponding textual contexts. 

\subsection{Mixed Augmentation on Synthetic-Aware Branch}
\label{3.3}
The key idea of this branch is to employ a combination of data augmentation techniques to mix synthetic data with real data, which guarantees that the model treats all data as same. To do so, we select two simple and efficient mix methods as follows.

\subsubsection{MixUp.}
First, we apply the MixUp \cite{zhang2017mixup} to mix real samples and synthetic samples. Let $x \in \mathbb{R}^{W \times H \times C}$ and $y$ denote a training image and its corresponding label, respectively. For a pair of two images and their labels $(x_{i}, y_{i})$ and $(x_{j}, y_{j})$, the mixed augmentation image and its label $(\tilde{x}_{ij}, \tilde{y}_{ij})$ can be calculated as follows:
\begin{equation}
\tilde{x}_{ij} = \lambda x_{i} + (1 - \lambda) x_{j}, \ \ \ \tilde{y}_{ij} = \lambda y_{i} + (1 - \lambda) y_{j}
\label{eq_1}
\end{equation}
where the combination ratio $\lambda$ is sampled from a $\text{Beta}$ distribution. $\lambda$ is a hyper-parameter, which is set as 1 here for a more random mixture.

\subsubsection{CutMix.}
Another mixture technology CutMix \cite{yun2019cutmix} also aims to mix real samples and synthetic samples, which combine two images by simply replacing the image region with a path from another image. Similarly, the combining operation can be defined as follows:
\begin{equation}
\tilde{x}_{ij} = \textbf{M} \odot x_{i} + (\textbf{1} - \textbf{M}) \odot x_{j}, \ \ \ \tilde{y}_{ij} = \lambda y_{i} + (1 - \lambda) y_{j}
\label{eq_2}
\end{equation}
where the matrix $\textbf{M} \in \{ 0, 1 \}^{W \times H}$ is the binary mask that indicates the randomly selected region from the image $x_{i}$, which is then filled into $x_{j}$. $\textbf{1}$ is a binary mask filled with ones. $\odot$ is element-wise multiplication. Specifically, we sample the bounding box coordinates $\textbf{B} = (r_{x}, r_{y}, r_{w}, r_{h})$ indicating the cropping regions on $x_{i}$ and $x_{j}$. The binary mask $\textbf{M}$ is determined by setting to 0 within the bounding box $\textbf{B}$, and to 1 elsewhere. The box coordinates are uniformly sampled according to:
\begin{equation} 
\begin{aligned}
r_{x} &\sim \text{Uniform}(0, W), \ \  r_{w} = W \sqrt{1 - \lambda} \\
r_{y} &\sim \text{Uniform}(0, H), \ \  r_{h} = H \sqrt{1 - \lambda}
\end{aligned}
\label{eq_3}
\end{equation}
where the $\lambda$ is also sampled from a $\text{Beta}$ distribution, ensuring that the cropped area ratio is equal to $1 - \lambda$.

After constructing two kinds of real-synthetic mixed augmented data, we use the cross-entropy loss to calculate two losses in the same way:
\begin{equation}
\begin{aligned}
\mathcal{L}_{MixUp} =& \ \mathcal{L}_{ce}(f(\tilde{x}_{MixUp}), \ \tilde{y}_{MixUp}) \\
\mathcal{L}_{CutMix} =& \ \mathcal{L}_{ce}(f(\tilde{x}_{CutMix}), \ \tilde{y}_{CutMix})
\label{eq_4}
\end{aligned}
\end{equation}
here, the $f(\tilde{x})$ denotes the predictions of $\tilde{x}$.

\subsection{Supervised Contrastive Learning on Synthetic-Aware Branch}
\label{3.4}
The main idea of the synthetic-aware branch is to enable the model to distinguish between real data and synthetic data, thereby learning their discrepancies and improving the model's understanding and generalization capabilities. 

\subsubsection{Preliminaries on Contrastive Learning.}
Contrastive learning methods typically use the noise contrastive estimation (NCE) objective to distinguish between similar and dissimilar data points. Specifically, it pulls different views of the same sample together, while pushing others away in a latent space. Similarly, in a supervised classification task, where label information is given, it pulls different views of samples from the same class together while pushing others away. For instance, we randomly sample a mini-batch of images $\{ \mathbf{x} \}^{N}_{i=1}$, and then we apply a random augmentation function $T(\cdot)$ to obtain a pair of different augmented views each image. Within this multiviewed batch, we can define that $i \in I \equiv \{1, 2, ..., 2N \}$ as the index of multiviewed batch samples. Following the previous works \cite{cui2021parametric,khosla2020supervised}, the loss function can be defined as:
\begin{equation}
    \label{eq_sc}
    \mathcal{L}_{SC} = \sum_{i \in I} \mathcal{L}_{SC}^{i} = \sum_{i \in I} \left( -\frac{1}{|I_y| - 1} \sum_{j \in \{I_y \} \backslash \{i\}} \log \frac{\exp(\mathbf{z}_i \cdot \mathbf{z}_j / \tau)}{\sum\limits_{k \in I \backslash \{i \} } \exp(\mathbf{z}_i \cdot \mathbf{z}_k / \tau)} \right)
\end{equation}
where $\mathbf{z}_i = \phi(\mathcal{F}(T(\mathbf{x}_i)))$ indicates the vector in a latent space of input images, and $I_y \equiv \{i \in I: \mathbf{y}_i = y \}$ indicates a set of views of same class $y$, and the symbol $|\cdot|$ denotes to the number of the elements, and $\tau > 0$ denotes a scalar temperature hyper-parameter.

\subsubsection{Label Correction.}
To further address the problem of low-quality synthetic images, motivated by \cite{zheng2021weakly}, we propose a K-Nearest-Neighbour(KNN) based method to dynamically detect low-quality synthetic image-label pairs. Specifically, given two sets of representations $\{ \mathbf{z}_{real} \}^{N_1}_{i=1}$, $\{ \mathbf{z}_{syn} \}^{N_2}_{i=1}$ and their corresponding labels $\{ \mathbf{y}_{real} \}^{N_1}_{i=1}$, $\{ \mathbf{y}_{syn} \}^{N_2}_{i=1}$ of both real images and synthetic images, we generate correction labels for the synthetic images based on the labels of their k nearest real neighbors. Combining them with original labels $\mathbf{y}_{org}$, the final labels for synthetic images can be defined as:
\begin{equation}
    \mathbf{y}_{new} = \left\{ \begin{array}{ll}
    \mathbf{y}_{org} & , \ \text{if} \ \ \ (\mathbf{y}_{cor}^{1} = \mathbf{y}_{org}) \ \& \ (\mathbf{y}_{cor}^{2} = \mathbf{y}_{org}) \\
    \mathbf{y}_{noise} & , \ \text{otherwise}
    \end{array} \right.
\end{equation}
where $\mathbf{y}_{new}$ indicates the new labels for the synthetic data. $\mathbf{y}_{cor}^{1}$ and $\mathbf{y}_{cor}^{2}$ are the two sets of corrective labels of two different augmented views. $\mathbf{y}_{noise}$ represents the noise labels for those low-quality synthetic images.
\begin{figure}
\centering
\includegraphics[width=0.8\textwidth]{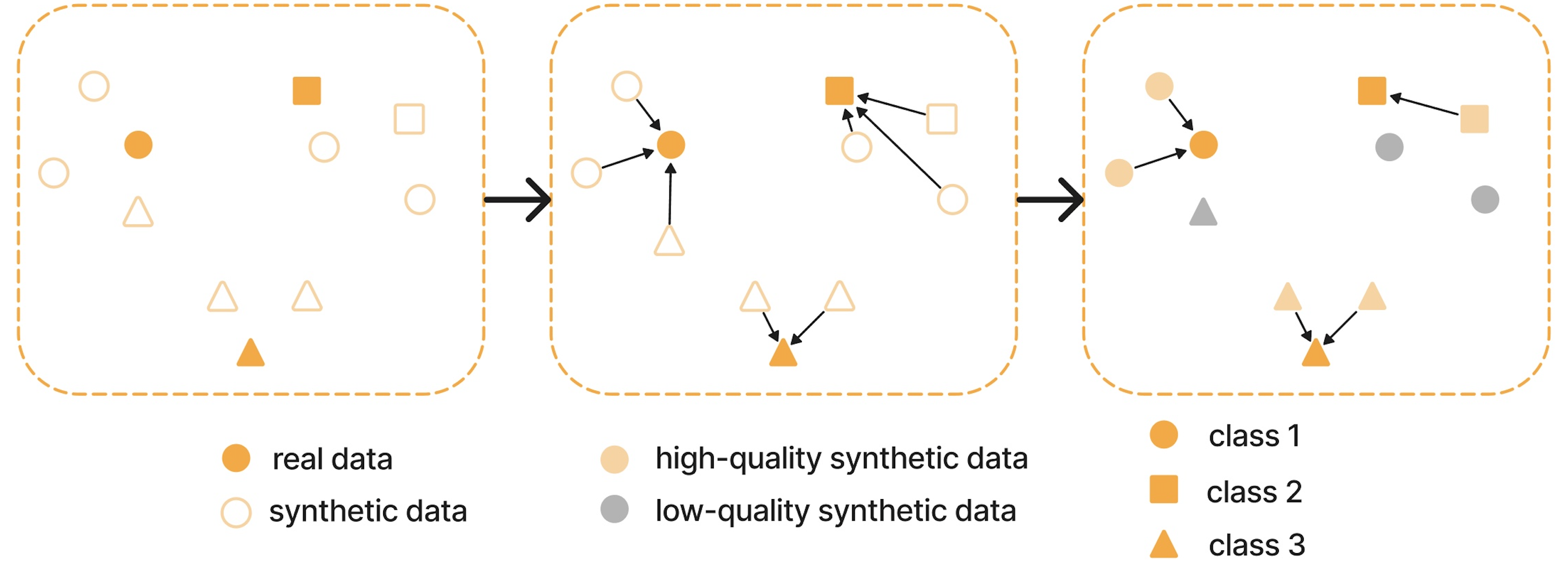}
\caption{An example of label correction process.} \label{label-correction}
\end{figure}

\subsubsection{Prototypes Complement.}
To ensure all classes of real samples appear in every mini-batch, following previous work \cite{nassar2023protocon}, we calculate prototypes of each class by averaging the real training data at the end of the previous epoch for the current epoch. Formally, the normalized prototype for a certain class $y$ can be formulated as: 
\begin{equation}
    \mathcal{\bar{P}}_y = \frac{1}{N_1} \sum_{\mathbf{z}_i \in \mathbf{z}_{real}}\mathbf{z}_i, \ \ \ \ \ \mathcal{P}_y = \frac{\mathcal{\bar{P}}_y}{{\lVert \mathcal{\bar{P}}_y \rVert}_2 }
\end{equation}
where $\mathcal{P} \in \mathbb{R}^{C \times d}$ indicates that prototypes include $C$ data points with $d$ dimension each. $N_1$ is the number of real data, and the symbol ${\lVert \cdot \rVert}_2$ denotes the $l^2$-norm. Note that since the prototypes $\mathcal{P}$ are the average from real data, we treat them as real data in the following declaration.

\subsubsection{Noise Dropping Strategy.}
The key idea to tackle low-quality synthetic data is treating them as noise within a mini-batch. Here, we propose three forms of loss for this scenario. 

First, the most intuitive way is simply to ignore all the noise data points: 
\begin{equation}
    \label{eq_l1}
    \mathcal{L}_{1} = \sum_{i \in A \backslash I^{'}} \left( -\frac{1}{|I_y|} \sum_{j \in A_y \backslash \{I^{'}_y \cup \{i\} \} } \log \frac{\exp(\mathbf{z}_i \cdot \mathbf{z}_j / \tau)}{\sum\limits_{k \in \{A_y \} \backslash \{I^{'}_y \cup \{i\} \} } \exp(\mathbf{z}_i \cdot \mathbf{z}_k / \tau)} \right)
\end{equation}
here, $A = \{I, I^{'}\}$ denotes the entire dataset, where $I$ consists of real data, prototypes, and high-quality synthetic data, and $I^{'}$ is the noise data. The symbol $|\cdot|$ represents the number of elements. Note the original $|I_y| - 1$ term in Eq. \ref{eq_sc} should plus one because prototypes contain one data point for each class. 

Second, instead of simply removing all the noise data points, we keep them and treat all the noise data points as individual positive samples. Concretely, we assign unique noise labels $\mathbf{y}_{noise} = \{-1, -2, ..., -N_n\}$ for noise data points, representing new classes distinct from all the original labels:
\begin{equation}
    \label{eq_l2}
    \mathcal{L}_{2} = \sum_{i \in A} \mathcal{L}_{2}^{i} = \sum_{i \in A} \left( -\frac{1}{|I_y| + |I^{'}_y|} \sum_{j \in A \backslash \{i\}} \log \frac{\exp(\mathbf{z}_i \cdot \mathbf{z}_j / \tau)}{\sum\limits_{k \in A \backslash \{i \} } \exp(\mathbf{z}_i \cdot \mathbf{z}_k / \tau)} \right)
\end{equation}
where $y_i \in \{\mathbf{y}_{org}, \ \mathbf{y}_{noise}\}$ indicates there is also a new noise class in addition to the original class. The denominator of the average term should count all samples among $I_y$ and $I^{'}_y$. 

Third, similar to $\mathcal{L}_{2}$, we keep noise data points, but only as negative samples:
\begin{equation}
    \label{eq_l3}
    \mathcal{L}_{3} = \sum_{i \in A \backslash I^{'}} \left( -\frac{1}{|I_y|} \sum_{j \in A_y \backslash \{I^{'}_y \cup \{i\} \}} \log \frac{\exp(\mathbf{z}_i \cdot \mathbf{z}_j / \tau)}{\sum\limits_{k \in A \backslash \{i \} } \exp(\mathbf{z}_i \cdot \mathbf{z}_k / \tau)} \right)
\end{equation}

In summary, the overall training loss function for our supervised contrastive learning framework can be expressed as: 
\begin{equation}
    \label{eq_l_overall}
    \mathcal{L}_{overall} = \lambda \mathcal{L}_{MixUp} + \beta \mathcal{L}_{CutMix} + \gamma \mathcal{L}_{1, 2, 3}
\end{equation}
where $\lambda$, $\beta$, $\gamma$ are the hyper-parameters. More details in Algorithm \ref{algo2}.

\begin{algorithm}
\caption{Learining algorithm}
\begin{algorithmic}[1]
\State \textbf{Input:} $\{ \mathbf{x}_{1} \}^{N}_{i=1}$ and $\{ \mathbf{x}_{2} \}^{N}_{i=1}$: two batches of samples.
$\{ \mathbf{y}_{1} \}^{N}_{i=1}$ and $\{ \mathbf{y}_{2} \}^{N}_{i=1}$: labels of samples.
$\mathbbm{1}$: indicator of whether the samples are synthetic. 
$T(\cdot)$: the augmentation function. 
$\mathcal{F}$: the backbone network. 
$classifer(\cdot)$: the classifier head of synthetic-unaware branch. 
$\phi$: the projection head of the synthetic-aware branch. 
\While{network not converge}
    \State Calculate class prototypes $\mathbf{p}^{2}$, $\mathbf{p}^{3}$;
    \For{$i = 1$ to step}
        \State $\mathbf{v}^{1}_{1}$, $\mathbf{v}^{2}_{1}$, $\mathbf{v}^{3}_{1} = T(\{ \mathbf{x}_{1} \}^{N}_{i=1})$
        \State \# Synthetic-unaware Branch
        \State $\mathbf{v}^{1}_{2}, \_, \_ = T(\{ \mathbf{x}_{2} \}^{N}_{i=1})$
        \State $\tilde{\mathbf{x}}_{m}$, $\tilde{\mathbf{y}}_{m} = MixUp(\mathbf{v}^{1}_{1}, \mathbf{v}^{1}_{2})$ Eq. \ref{eq_1}; \ \ $\tilde{\mathbf{x}}_{c}$, $\tilde{\mathbf{y}}_{c} = CutMix(\mathbf{v}^{1}_{1}, \mathbf{v}^{1}_{2})$ Eq. \ref{eq_2}
        \State $\mathbf{h}_{m} = \mathcal{F}(\tilde{\mathbf{x}}_{m})$, $\mathbf{h}_{c} = \mathcal{F}(\tilde{\mathbf{x}}_{c})$
        \State $\hat{\mathbf{y}}_{m} = classifer(\mathbf{h}_{m})$, $\hat{\mathbf{y}}_{c} = classifer(\mathbf{h}_{c})$
        \State Calculate $\mathcal{L}_{MixUp}(\hat{\mathbf{y}}_{m}, \tilde{\mathbf{y}}_{m})$ and $\mathcal{L}_{CutMix}(\hat{\mathbf{y}}_{c}, \tilde{\mathbf{y}}_{c})$ Eq. \ref{eq_4}
        \State \# Synthetic-aware Branch
        \State $\mathbf{h}^{2} = \mathcal{F}(\mathbf{v}_{1}^{2})$, $\mathbf{h}^{3} = \mathcal{F}(\mathbf{v}_{1}^{3})$
        \State $\mathbf{z}^{2} = \phi(\mathbf{h}^{2})$, $\mathbf{z}^{3} = \phi(\mathbf{h}^{3})$
        \State Generate new labels $\mathbf{y}_{1}^{2}$, $\mathbf{y}_{1}^{3}$ for $\{ \mathbf{x}_{1} \}^{N}_{i=1}$ based on $\mathbf{z}^{2}$, $\mathbf{z}^{3}$.
        \State Calculate $\mathcal{L}_{SC}(\mathbf{z}^{2}, \mathbf{z}^{3}, \mathbf{p}^{2}, \mathbf{p}^{3}, \mathbf{y}_{1}^{2}, \mathbf{y}_{1}^{3})$ Eq. \ref{eq_l1}, \ref{eq_l2}, \ref{eq_l3}
        \State Optimize the network by $\mathcal{L}_{overall}$ Eq. \ref{eq_l_overall}
    \EndFor
\EndWhile
\State \textbf{Output:} The well trained model $\mathcal{F}$ and $classifer(\cdot)$
\end{algorithmic} \label{algo2}
\end{algorithm}

\section{Experiments}
\subsection{Experiment Setup}
\subsubsection{Datasets.}
Both CIFAR-10 and CIFAR-100 \cite{krizhevsky2009learning} consist of 60000 32$\times$32 color images with 50000 images for training and 10000 images for testing. There are 10 classes and 100 classes resulting in 6000 images and 600 images for each class respectively. CIFAR-10-LT and CIFAR-100-LT are the subsets of the long-tailed versions of CIFAR-10 and CIFAR-100. In this work, the training datasets consist of real data and generated data. For real data, following the previous works \cite{alshammari2022long,du2023global,zhu2022balanced}, we use the exponential decay function $N_{i} = N_{o}\mu ^ { \frac{i}{n - 1} }$ to obtain long-tailed datasets, where $i$ is the class index (0-indexed), $N_{o}$ is the original number of training images and $\mu$ is the imbalanced factor. The imbalanced factor $\mu = N_{max} / N_{min}$ reflects the degree of imbalance and we use different $\mu$ [10, 50, 100, 200] for both CIFAR-10-LT and CIFAR-100-LT. For synthetic data, they are simply set as the complement of the real data. Following the official GLIDE T2I process, we use different prompts to generate 64$\times$64 color synthetic images.

ImageNet-LT, proposed by \cite{liu2019large}, is a long-tailed version of the original ImageNet \cite{deng2009imagenet} dataset, which is designed to mimic real-world scenarios. Specifically, the original ImageNet consists of 1,280,000 images of 1000 classes in total with 1280 images per class. And ImageNet-LT is a subset of ImageNet with an imbalance factor of 256. To reduce computational costs, we only use basic prompts to generate 256$\times$256 color synthetic images.

\subsubsection{Implementation Details}
We use Pytorch \cite{paszke2019pytorch} deep learning framework to train models for all datasets. 

For CIFAR-10-LT and CIFAR-100-LT, we follow \cite{du2023global,cui2021parametric,zhu2022balanced} to use ResNet-32 \cite{he2016deep} for a fair comparison across different studies. For the optimization, we use a standard SGD optimizer with a momentum of 0.9. We start with an initial learning rate of 0.01 with a cosine annealing scheduler. Additionally, the batch size is set at 128 and the weight decay rate at 5e-3. The model will be trained for 200 epochs, after which the model is evaluated on the test set to assess its performance. In terms of data augmentation strategies, we use random horizontal flipping and cropping for the classification branch, AutoAugment \cite{cubuk2019autoaugment}, Cutout \cite{devries2017improved} and SimAugment \cite{chen2020simple} for contrastive learning branch. And MLP consists of two hidden layers of size 2048 and output vector of size 128, and there are batch normalization and ReLU activation function between two hidden layers. 

For ImageNet-LT, we use ResNet-50 \cite{he2016deep} and ResNeXt-50-32x4d \cite{xie2017aggregated} as backbone. We use a standard SGD optimizer with a momentum of 0.9. We set the initial learning rate as 0.1, which is adjusted by a cosine annealing scheduler. The batch size is set at 256 and the weight decay rate at 1e-4. The model is trained over 180 epochs.

\subsubsection{Evaluation Protocol}
For all datasets, we focus on the top-1 accuracy. We train models on a balanced dataset composed of a mix of real long-tailed training sets and synthetic set and evaluate them on the balanced validation/test dataset. Furthermore, for ImageNet-LT, we report many-shot classes with real training samples > 100, medium-shot classes with real training samples 20 $\sim$ 100, and few-shot classes with real training samples $\le$ 20, respectively.

\begin{table}
  \caption{Top-1 accuracy of ResNet-32 on CIFAR-10-LT and CIFAR-100-LT with different imbalance factors [200, 100, 50, 10]. The best results are marked in bold. $\dagger$ denotes results reproduced by ourselves using the code released by authors. * reports the results based on real-synthetic mixed balanced data.}
  \label{cifar_result}
  \centering
  \begin{tabular}{lcccccccc}
    \hline
    & \multicolumn{4}{c}{CIFAR-10} & \multicolumn{4}{c}{CIFAR-100} \\
    \cmidrule(lr){2-5} \cmidrule(lr){6-9}
    Method & IF=200 & IF=100 & IF=50 & IF=10 & IF=200 & IF=100 & IF=50 & IF=10 \\
    \hline
    CE & 65.87 & 70.14 & 74.94 & 86.18 & 34.70 & 38.46 & 44.02 & 55.73  \\
    CB-Focal \cite{cui2019class} & 68.89 & 74.57 & 79.27 & 87.49 & 36.23 & 39.60 & 45.42 & 57.99 \\
    Mixup \cite{zhang2017mixup} & - & 73.06 & 77.82 & 87.10 & - & 39.54 & 54.99 & 58.02 \\
    PaCo \cite{cui2021parametric} & - & - & - & - & - & 52.0 & 56.0 & 64.2 \\
    ProCo \cite{du2024probabilistic} & - & 85.9 & 88.2 & 91.9 & - & 52.8 & 57.1 & 65.5 \\
    Hybrid-SC \cite{wang2021contrastive} & - & 81.40 & 85.36 & 91.12 & - & 46.72 & 51.87 & 63.05 \\
    BCL \cite{zhu2022balanced} & - & 84.32 & 87.24 & 91.12 & - & 51.93 & 56.59 & 64.87  \\
    GLMC$^{\dagger}$ \cite{du2023global} & - & 87.25 & 90.39 & 94.33 & - & 57.70 & 62.70 & 72.63 \\
    SURE \cite{li2024sure} & - & 86.93 & 90.22 & 94.96 & - & 57.34 & 63.13 & 73.24 \\
    \midrule
    CE$^{*}$ & 82.29 & 84.45 & 87.27 & 91.10 & 55.24 & 57.80 & 59.85 & 67.93 \\
    GLMC$^{*}$ & 85.13 & 88.33 & 91.35 & 94.80 & 58.46 & 61.80 & 65.21 & 74.11 \\
    SAU$^{*}$ (Ours) & \textbf{89.96} & \textbf{92.21} & \textbf{93.91} & \textbf{95.88} & \textbf{61.95} & \textbf{64.47} & \textbf{68.22} & \textbf{76.31} \\
    \bottomrule
  \end{tabular}
\end{table} 

\begin{table}
    \caption{Top-1 accuracy (\%) of on ImageNet-LT dataset compared to the state-of-the-art works with different backbone. The best results are marked in bold. * reports the results based on real-synthetic mixed balanced data.}
    \label{imagenet_result}
    \centering
    \begin{tabular}{lccccc}
    \toprule
    Method & Backbone & Many  & Med   & Few   & All   \\ 
    \midrule
    CE & ResNet-50 & 64 & 33.8  & 5.8 & 41.6  \\
    CB-Focal \cite{cui2019class} & ResNet-50   & 39.6  & 32.7  & 16.8  & 33.2  \\
    BCL (90 epochs) \cite{zhu2022balanced}      & ResNeXt-50  & 67.2  & 53.9  & 36.5  & 56.7  \\
    BCL (180 epochs) \cite{zhu2022balanced}     & ResNeXt-50  & 67.9  & 54.2  & 36.6  & 57.1  \\
    PaCo (400 epochs) \cite{cui2021parametric} & ResNeXt-50   & -  & -  & -  & \textbf{58.2}  \\
    PaCo (180 epochs) \cite{cui2021parametric} & ResNeXt-50 & 64.4 & \textbf{55.7} & 33.7 & 56.0 \\
    ProCo (180 epochs) \cite{du2024probabilistic} & ResNet-50  & 68.2 & 55.1 & 38.1 & 57.8 \\
    GLMC \cite{du2023global} & ResNeXt-50  & \textbf{70.1}  & 52.4  & 30.4  & 56.3  \\
    \midrule
    CE$^{*}$  & ResNet-50  & 58.3 & 47.7 & 32.1 & 49.7 \\
    SAU$^{*}$ (ours) & ResNet-50  & 62.3 & 51.7 & 37.2 & 53.7 \\
    SAU$^{*}$ (ours) & ResNeXt-50  & 64.7 & 52.6 & \textbf{38.4} & 55.2 \\
    \bottomrule
    \end{tabular}
\end{table}

\subsection{Main Results}
\subsubsection{CIFAR-10-LT and CIFAR-100-LT}
The result has been reported in Table \ref{cifar_result}. For comparison baselines , we mainly consider previous state-of-the-art works including CB-Focal \cite{cui2019class}, PaCo \cite{cui2021parametric}, Hybrid-SC \cite{wang2021contrastive}, BCL \cite{zhu2022balanced}, GLMC \cite{du2023global}, and SURE \cite{li2024sure}, whose models are trained from scratch. Furthermore, for a fairer comparison, we also apply recent state-of-the-art work GLMC on real-synthetic mixed balanced datasets. As we can see, our method achieves the best results on all imbalance factors.

\subsubsection{ImageNet-LT}
Similar to CIFAR, we report the Top-1 accuracy of our models compared to other works in Table \ref{imagenet_result}. For comparison baselines, we mainly consider both current and previous state-of-the-art works including Cross-Entropy, CB-Focal \cite{cui2019class}, BCL \cite{zhu2022balanced}, PaCo \cite{cui2021parametric}, ProCo \cite{du2024probabilistic} and GLMC \cite{du2023global}. From the results, our method achieves the best performance in few-shot.

\begin{table}
  \caption{Effectiveness of different noise dropping strategies. We report the Top-1 accuracy on CIFAR-10-LT and CIFAR-100-LT (IF=100) with the ResNet-32 backbone. }
  \label{loss_compare}
  \centering
  \begin{tabular}{ccc}
    \toprule
    Method & CIFAR-10-LT & CIFAR-100-LT \\
    \midrule
    $\mathcal{L}_{1}$ & 91.34 & 63.03 \\
    $\mathcal{L}_{2}$ & \textbf{92.21} & \textbf{64.47} \\
    $\mathcal{L}_{3}$ & 91.05 & 64.28 \\
    \bottomrule
  \end{tabular}
\end{table}

\begin{table}
  \caption{Effectiveness of primary components. We report Top-1 Acc. on CIFAR-100-LT (IF=100). The backbone is ResNet-32. Note that we use $\mathcal{L}_{2}$ as $\mathcal{L}_{SC}$ for all experiments. }
  \label{component}
  \centering
  \begin{tabular}{ccccc}
    \toprule
    $\mathcal{L}_{CE}$& $\mathcal{L}_{MixUp}$ & $\mathcal{L}_{CutMix}$ & $\mathcal{L}_{SC}$ & Top-1 Acc. \\
    \midrule
    $\checkmark$ & & & & 57.84 \\
    \midrule
    & $\checkmark$ & & & 59.95 \\
    & & $\checkmark$ & & 58.22 \\
    & $\checkmark$ & $\checkmark$ & & 61.68 \\
    \midrule
    & $\checkmark$ &  & $\checkmark$ & 62.45 \\
    & & $\checkmark$ & $\checkmark$ & 61.09 \\
    \midrule
    & $\checkmark$ & $\checkmark$ & $\checkmark$ & \textbf{64.47} \\
    \bottomrule
  \end{tabular}
\end{table}

\subsection{Ablation Studies}
\subsubsection{Effectiveness of Noise-Dropping Strategy.}
First, we compare the performances of three different contrastive losses ($\mathcal{L}_{1}$, $\mathcal{L}_{2}$, and $\mathcal{L}_{3}$) to deal with the poor-quality synthetic images. From the results presented in Table \ref{loss_compare}, all losses are effective and $\mathcal{L}_{2}$ performs best on both CIFAR-10-LT and CIFAR-100-LT. 

\subsubsection{Effectiveness of Components.}
We also deploy an extensive ablation experiment to examine the effectiveness of primary components. The results are shown in Table \ref{component}, which validates the effectiveness of our proposed method. Note the first row is the baseline, which uses cross-entropy as loss function and is trained on the real-synthetic mixed dataset. As we can see, the final row of the table, where both $\mathcal{L}_{MixUp}$ and $\mathcal{L}_{CutMix}$ are utilized alongside $\mathcal{L}_{SC}$, presents the best result with a Top-1 accuracy of \textbf{64.47\%}.

\section{Conclusion}
In this work, to enhance the long-tailed image recognition, we leverage LLMs and T2I models to generate synthetic images, which is used as a complement to the long-tailed distribution dataset to obtain a balanced dataset. And we propose a powerful framework SAU to tackle such real-synthetic mixed datasets. 
As a result, SAU has achieved state-of-the-art Top-1 accuracy on both CIFAR-10-LT and CIFAR-100-LT, and significantly improve the performance of tail classes on ImageNet-LT.

%
%
%
\bibliographystyle{splncs04}
\bibliography{references}

\end{document}